# Recurrent Neural Networks Beyond Time: Learning from Multiple Ordered Projections

Vagan Terziyan [1], Artur Terziian [2], Oleksandra Vitko [3]

[1] *Faculty of Information Technology, University of Jyväskylä, FINLAND; (Contact author: vagan.terziyan@jyu.fi; ORCID: 0000-0001-7732-2962)*

[2] *Faculty of Informatics and Statistics, Prague University of Economics and Business, Czech Republic; tera03@vse.cz*

[3] *Department of Artificial Intelligence, Kharkiv National University of Radio Electronics, Ukraine; oleksandra.vitko@nure.ua*

## Abstract

Recurrent neural networks (RNNs) are widely used for sequence learning, yet their application is commonly associated with temporal data, although recurrent computation fundamentally operates on ordered sequences rather than on time itself. Building on this observation, we introduce the Ordered Structural Dependency Hypothesis (OSDH), which proposes that multiple admissible orderings of the same observations may reveal complementary structural dependencies inaccessible through a single sequential organization. To operationalize this hypothesis, we propose the Independent Structural Expert Principle (ISEP), whereby projection-specific sequence models are trained independently before their learned representations are integrated through a dedicated fusion model. As a concrete realization, we present Structural Evolution RNNs (SE-RNNs), which employ conventional RNNs as projection-specific structural experts while preserving the underlying recurrent computation unchanged. Proof-of-concept experiments on three synthetic datasets with substantially different levels of structural complexity demonstrate that the proposed architecture consistently benefits from multiple ordered projections when hidden structural dependencies are present, while remaining competitive on simpler datasets. Since OSDH is independent of the underlying sequence-processing model, the proposed framework naturally extends beyond recurrent networks and may be instantiated using alternative architectures. The results suggest a general computational perspective for exploiting complementary ordered representations across diverse structured learning problems.

## 1. Introduction

Artificial intelligence (AI) seeks to construct computational models capable of understanding, predicting, and explaining the behavior of complex real-world systems. Such systems are naturally described by multiple interacting variables, each representing a particular aspect of the observed world. Learning from these systems ultimately requires discovering hidden statistical regularities embedded within the observed data. Over the past decades, recurrent neural networks (RNNs) and their gated variants have become one of the principal computational tools for this purpose, achieving remarkable success in applications ranging from natural language processing and speech recognition to financial forecasting, robotics, industrial control, and scientific data analysis [1-3]. In almost all these applications, however, recurrent learning is driven by a single underlying assumption: observations are first ordered according to *time*, and the recurrent model is then expected to discover temporal dependencies within the resulting sequence.

This temporal interpretation has become so dominant that recurrent neural computation is often regarded as being inherently tied to time-series analysis. Nevertheless, the recurrence mechanism itself possesses no intrinsic notion of physical time. Mathematically, an RNN simply receives an ordered sequence of observations and recursively propagates information from preceding elements to subsequent ones. Whether the ordering represents chronological time, spatial position, or another ordered attribute is not determined by the recurrent operator itself but by the way the input sequence is constructed. In other words, recurrence is fundamentally an operator over ordered observations, while temporal interpretation is imposed externally through data representation.

In many real-world problems, data exhibit rich internal structure that cannot be adequately captured by a single privileged ordering. Spatial configurations, attribute-wise relationships, semantic groupings, or domain-specific decompositions often induce multiple meaningful orderings over the same underlying object. Classical neural architectures typically collapse these structures into a single vector representation (as in feedforward networks) or impose a single sequence (as in RNNs), thereby discarding potentially valuable contextual information [4]. While multi-dimensional recurrent models and attention-based mechanisms have attempted to address aspects of this limitation, they usually rely on tightly coupled recurrence over structured grids or on global context aggregation, rather than explicitly reasoning over multiple independent ordered projections [5-6].

This observation raises a natural question. *Is temporal ordering fundamentally unique for recurrent learning, or is it merely one possible ordering among many that may expose useful statistical dependencies?* If recurrent computation depends only on the existence of an ordered sequence, then the conventional temporal sequence may represent only one particular view of a considerably richer learning problem.

Throughout this work we regard the observed world $W$ as a multidimensional system, which has temporal coordinate $T$ ("time"), $X_1, \dots, X_m$ as other coordinates describing the same observations, and some target attribute $Y$. Traditional recurrent learning constructs exactly one sequence by sorting available observations according to the temporal coordinate $T$. The central hypothesis of this paper is that other coordinates may also induce meaningful orderings of the same observations and therefore reveal complementary dependency patterns that remain hidden under temporal ordering alone.

To formalize this idea, we introduce the Ordered Structural Dependency Hypothesis (OSDH) as follows. Let $X_i$ be any coordinate that defines meaningful ordering of observations. After sorting the observations according to $X_i$ (using a deterministic tie-breaking rule when necessary), the resulting sequence may expose sequential statistical dependency patterns among neighboring observations that are not observable under the conventional temporal ordering. RNNs can exploit such dependencies regardless of whether the ordering corresponds to physical time or another ordered coordinate.

OSDH should not be interpreted as claiming that every variable of a dataset is suitable for recurrent processing. Instead, it states that *every coordinate capable of inducing a meaningful ordering of observations becomes a potential candidate for recurrent modeling.* The conditions under which a coordinate is considered admissible are developed formally in Section 3. Under this viewpoint, temporal dependency becomes one particular instance of a broader family of ordered structural dependencies rather than the defining object of recurrent learning itself.

The proposed conceptual shift is illustrated in **Figure 1**. Here traditional recurrent learning constructs a single temporal sequence by ordering observations according to time. Under the proposed framework, the same observations are independently reordered according to $M$ admissible coordinates, producing multiple ordered projections. Each projection is processed by an independent recurrent learner, after which the learned representations are integrated into a unified predictive model.

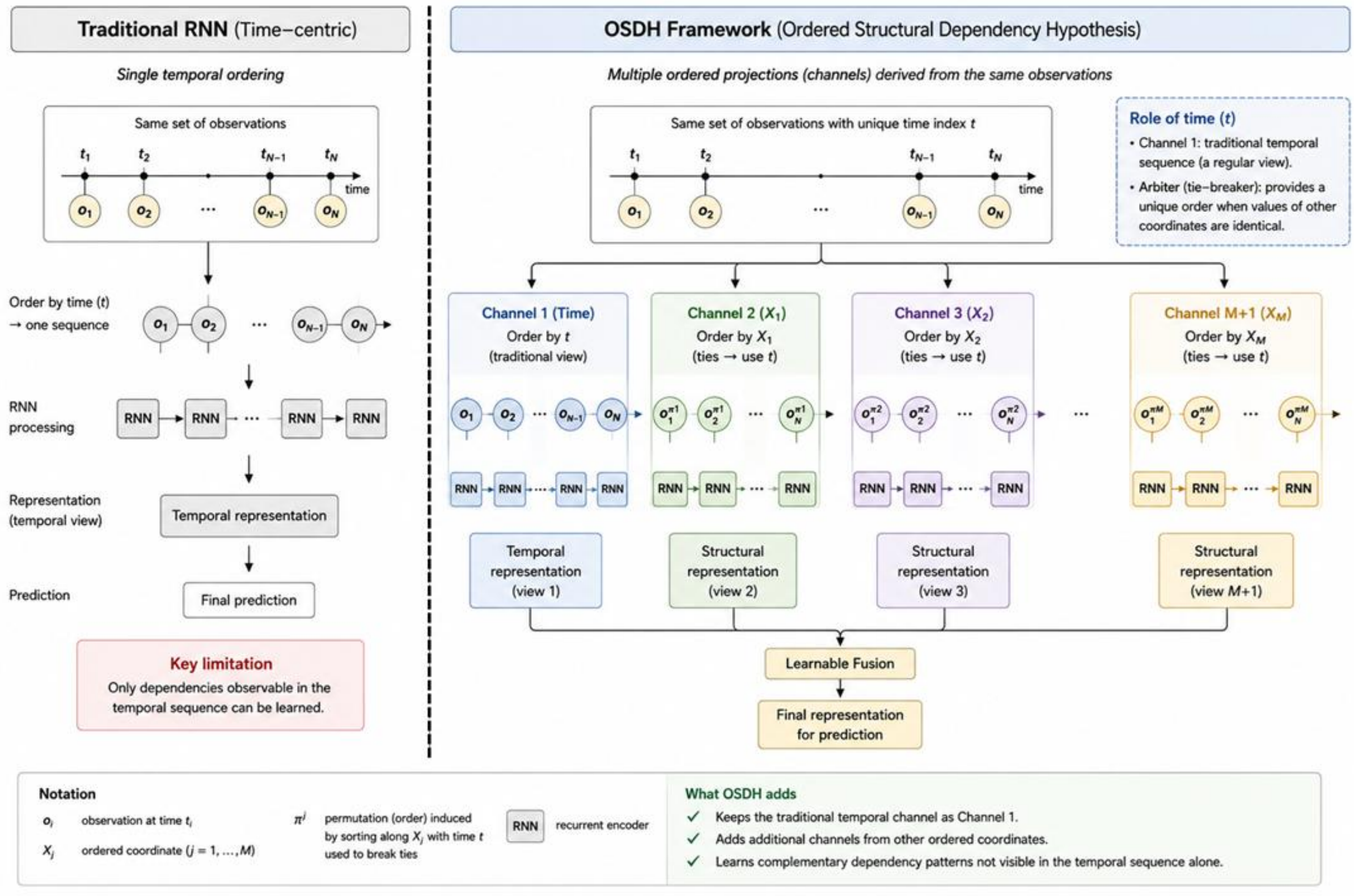

**Figure 1.** *From temporal recurrence to ordered structural recurrence*: Conventional RNNs learn from a single temporal ordering of observations. The proposed Ordered Structural Dependency Hypothesis extends this view by treating time as one member of a broader family of admissible ordered coordinates, each defining an independent recurrent learning channel. The temporal projection is preserved as the first channel, while time simultaneously serves as a canonical index for constructing deterministic non-temporal projections.

The key idea is therefore not to construct different datasets but to construct different ordered views of the same dataset. Each admissible coordinate defines its own ordering over exactly the same observations. Sorting observations according to different coordinates produces different sequential representations, each potentially exposing structural regularities invisible within the others. These ordered projections are complementary rather than competing, allowing recurrent models to extract diverse forms of structural information from the same underlying data.

An important distinction should be emphasized. Within the proposed OSDH framework, time plays a dual role. First, it is treated as an ordinary ordered coordinate that defines the conventional temporal recurrent channel, alongside all other admissible ordered coordinates. From the perspective of recurrent learning, no coordinate is inherently privileged: each ordered projection is processed by the same recurrent operator under the same computational principles. Second, time serves as the canonical observation index during data preparation. Every observation is assumed to possess a unique temporal index, which provides an unambiguous identifier of observations and a deterministic mechanism for resolving ordering ambiguities whenever identical values occur in other ordered coordinates during projection construction. Consequently, time is not privileged by the learning model itself, but only by the preprocessing stage, where it guarantees the deterministic construction of ordered projections while preserving a unique correspondence between the original observations and their multiple structural representations.

The intuition behind OSDH can be illustrated using chess. A chess game is conventionally represented as a temporal sequence of board states ordered by move number [7]. However, the same sequence of positions may also be reordered according to other naturally ordered coordinates of the board, such as files or ranks. Each of the orderings generates a different sequence containing the same observations but arranged according to a different structural principle. Neighboring positions within these alternative sequences may exhibit statistical regularities that differ from those observed under temporal ordering. Although such representations appear unconventional from a human perspective, they satisfy precisely the same computational requirement imposed by RNNs: they form ordered sequences of observations. This motivating example is developed in detail in the following section before the general mathematical framework is introduced.

The proposed hypothesis should also be distinguished from the well-known observation that RNNs can process arbitrary sequences, including DNA or protein sequences, symbolic strings, or other non-temporal data. This capability has been recognized for decades. Our contribution is fundamentally different. We do not merely observe that RNNs can process non-temporal sequences; rather, we propose that *ordered structural dependency itself should be regarded as the primary computational object of recurrent learning*, while temporal dependency represents only one particular realization of this broader concept. Consequently, the central design question shifts from traditional "*Can recurrent networks process non-temporal sequences?*" to "*Which ordered projections of an arbitrary multidimensional system reveal informative structural dependencies suitable for recurrent learning?*".

Based on the proposed OSDH framework, we introduce Structural Evolution RNNs (SE-RNNs) as one concrete realization of this broader computational viewpoint. The principal contribution of this work is not a new recurrent cell or a modification of existing recurrent architectures. Rather, it is a reformulation of what recurrent computation is fundamentally applied to. The proposed framework remains grounded in the well-established formalism of recurrent networks and ordered sequence processing [8], while complementing broader research directions in multi-view learning, modular neural systems, and representation learning [9]. Although this paper adopts RNNs as the underlying sequence-processing mechanism, the OSDH is not inherently tied to a particular neural architecture. In principle, any model capable of learning from ordered sequences, including more recent sequence-processing architectures such as Transformers [6], could potentially operate on the ordered projections defined by OSDH.

The present work intentionally focuses on recurrent networks because they provide the most direct and mathematically transparent realization of the proposed hypothesis. Within this framework, the original dataset is projected onto multiple admissible ordered coordinates. The conventional temporal projection is preserved as one projection-specific recurrent expert (or RNN channel as shown in **Figure 1**), while additional experts operate on ordered projections induced by other admissible coordinates. Each ordered projection is processed independently by its own recurrent encoder, allowing complementary structural dependency patterns to be learned in parallel. The resulting representations are subsequently integrated with a conventional feedforward backbone through a learnable fusion mechanism. OSDH should therefore be understood as an architecture-independent computational principle, whereas SE-RNN constitutes its first concrete realization using RNNs. Future implementations may replace the recurrent experts with alternative sequence-processing architectures while preserving the theoretical foundations of OSDH.

The main contributions of this work are summarized as follows:

- As a conceptual contribution, we formulate the OSDH, proposing that recurrent computation should be viewed as an operator over ordered structural dependencies rather than as a mechanism inherently associated with temporal evolution. Temporal dependency becomes one particular instance of a broader computational principle.
- As a methodological contribution, we develop a general framework for transforming a multidimensional dataset into multiple ordered projections by sorting the same observations according to different admissible coordinates. This establishes a systematic preprocessing pipeline enabling multi-channel recurrent learning.
- As an architectural contribution, we introduce SE-RNNs, a modular architecture in which independent recurrent branches learn complementary ordered structural dependencies before their representations are integrated through a trainable fusion module.
- As an experimental contribution, we present a proof-of-concept empirical evaluation on synthetic datasets exhibiting progressively richer hidden structural relationships. The experiments provide initial evidence that exploiting multiple ordered structural projections can improve predictive performance while component-wise analyses clarify the contributions of the feedforward backbone, projection-specific recurrent branches, and the fusion mechanism.

More broadly, this work advocates a different way of thinking about recurrent learning. Rather than treating time as the unique dimension deserving recurrent analysis, we argue that recurrence should be understood as a general computational operator over ordered structural dependencies. Under this interpretation, temporal modeling becomes one important (but not exclusive!) instance of a broader computational paradigm. We believe that this perspective opens new directions for designing neural architectures capable of integrating complementary structural views of complex multidimensional systems while remaining fully compatible with the established mathematical formalism of RNNs.

The remainder of this paper is organized as follows. Section 2 introduces the motivating chess example and illustrates the intuition behind multiple ordered structural projections. Section 3 develops the mathematical foundations of OSDH, formally defines admissible ordered coordinates, describes the projection procedure that transforms a multidimensional dataset into multiple ordered sequences, and discusses practical issues including repeated coordinate values and deterministic tie resolution. Section 4 presents the SE-RNN architecture. Section 5 reports proof-of-concept experimental results and analyzes the contribution of individual architectural components. Section 6 discusses related work and positions the proposed framework within the broader landscape of recurrent, multi-view, and modular neural architectures. Finally, Section 7 concludes the paper and outlines directions for future research. Supplementary Material, which is available online in: https://ai.it.jyu.fi/experiments/SE-RNNs/, provides full implementation details (source code, datasets, reproducibility artifacts, etc.).

## 2. Motivating Example: Multi-Axis Chess Game Representation

To provide an intuitive illustration of the proposed OSDH, we consider the familiar domain of chess. The example is intentionally simple and highly structured, allowing the central idea of the proposed framework to be understood before introducing the general mathematical formulation developed in the next section.

A chess game is traditionally represented as a sequence of board positions ordered by move number. This temporal ordering naturally defines the input sequence processed by conventional RNNs, where each board state is interpreted as the consequence of preceding game states. In this representation, recurrent computation is performed exclusively along the temporal coordinate, and learning aims to capture dependencies between successive positions.

The OSDH perspective begins with a different observation. A chess game is not characterized solely by temporal evolution. Every board position simultaneously possesses several ordered coordinates that describe the same underlying game from different structural perspectives. Besides the temporal coordinate (move number), the board itself contains naturally ordered spatial coordinates, namely files (columns) and ranks (rows). Each of these coordinates induces its own meaningful ordering of observations and therefore provides an alternative view of the same game. Unlike many real-world datasets discussed later in this paper, the chess domain offers an exceptionally clean illustration of this idea. The spatial coordinates are discrete, naturally ordered, and independent of one another, making them ideal for explaining the concept of multiple ordered projections without introducing additional preprocessing considerations.

The central intuition of OSDH is therefore straightforward: instead of assuming that recurrent learning should operate only over the temporal ordering of observations, we ask whether recurrent computation may also exploit structural dependencies revealed when exactly the same observations are reorganized according to other admissible ordered coordinates.

**Figure 2** illustrates this idea conceptually. Instead of observing the game only as a temporal sequence,

$$\text{Board}(t_1) \rightarrow \text{Board}(t_2) \rightarrow \cdots \rightarrow \text{Board}(t_{\text{endgame}}),$$

the same collection of board positions can also be reorganized according to ordered spatial coordinates, producing independent file projections and rank projections. Each projection preserves the original observations while changing only the ordering under which they are presented to the learning algorithm.

From the OSDH perspective, these projections define three independent sequential views of the same underlying world:

- *Temporal projection*: The conventional representation in which board positions are ordered by move number. This projection captures the sequential evolution of the game and corresponds to the traditional application of RNNs.
- *File projection*: Board positions are reorganized according to board files (columns). Such projections may expose structural regularities related to vertical piece interactions, pawn structures, control of open files, or recurring positional motifs that are less apparent under purely temporal ordering.
- *Rank projection*: Board positions are reorganized according to board ranks (rows). These projections may reveal complementary structural regularities associated with horizontal coordination of pieces, defensive formations, or positional evolution along ranks.

Importantly, OSDH does not claim that file or rank projections necessarily contain useful predictive information. Rather, it states that whenever an ordered coordinate induces statistically meaningful dependencies among neighboring observations, recurrent computation may exploit these dependencies regardless of whether the ordering corresponds to physical time or to another ordered attribute.

This distinction is fundamental. The proposed framework does not redefine causality, nor does it suggest that spatial coordinates replace temporal dynamics. Instead, temporal dependency is regarded as one particular instance of a broader class of ordered structural dependencies that may exist along different coordinates describing the same phenomenon.

Throughout this example, move number (time) plays two conceptually distinct roles. First, it defines the conventional temporal projection that constitutes the first recurrent channel of the proposed framework. Second, it uniquely identifies each board position, making the temporal ordering the natural reference representation of the game. Because the ordered coordinates considered in this example (time, files, and ranks) are discrete and unambiguous, no additional mechanisms are required to construct the corresponding ordered projections. The more general case, where repeated coordinate values require deterministic tie resolution during data preparation, is addressed in the next section.

From a machine learning perspective, the three projections should not be interpreted as additional handcrafted features. Instead, they represent complementary structural views of the same observations. Each projection organizes identical data according to a different ordering and therefore may expose different statistical dependency patterns. If such complementary dependencies exist, independently trained recurrent models can specialize in learning them.

Consequently, rather than relying exclusively on a single temporal recurrent model, the proposed framework assigns an independent artificial recurrent learner to each admissible ordered projection. Each learner develops expertise with respect to its own structural view of the data, while a subsequent integration mechanism combines these complementary representations into a unified prediction. The architectural realization of this idea is presented later in the paper.

To illustrate the proposed concept, we consider the well-known Scholar's Mate, represented in Portable Game Notation (PGN):

1. e4 e5 2. Qh5 Nc6 3. Bc4 Nf6 4. Qxf7# 1-0 .

This game can be represented as a three-dimensional tensor $X \in \mathcal{P}^{8\times8\times T}$, where the first two dimensions correspond to board files and ranks, respectively, the third dimension represents move number, and the set of possible board cell contents is denoted as $\mathcal{P} = \{\emptyset,$ ♙, ♘, ♗, ♖, ♕, ♔, ♟, ♞, ♝, ♜, ♛, ♚$\}$.

**Figure 2** visualizes the same game through the three ordered projections discussed above. The top row shows the conventional temporal projection consisting of successive board positions. The middle row presents the corresponding file projections, while the bottom row shows the rank projections. Although these non-temporal representations appear unfamiliar from the perspective of traditional chess analysis, they preserve exactly the same observations while reorganizing them according to different ordered coordinates. Consequently, each projection may reveal structural regularities that remain less visible when learning is restricted to the temporal ordering alone.

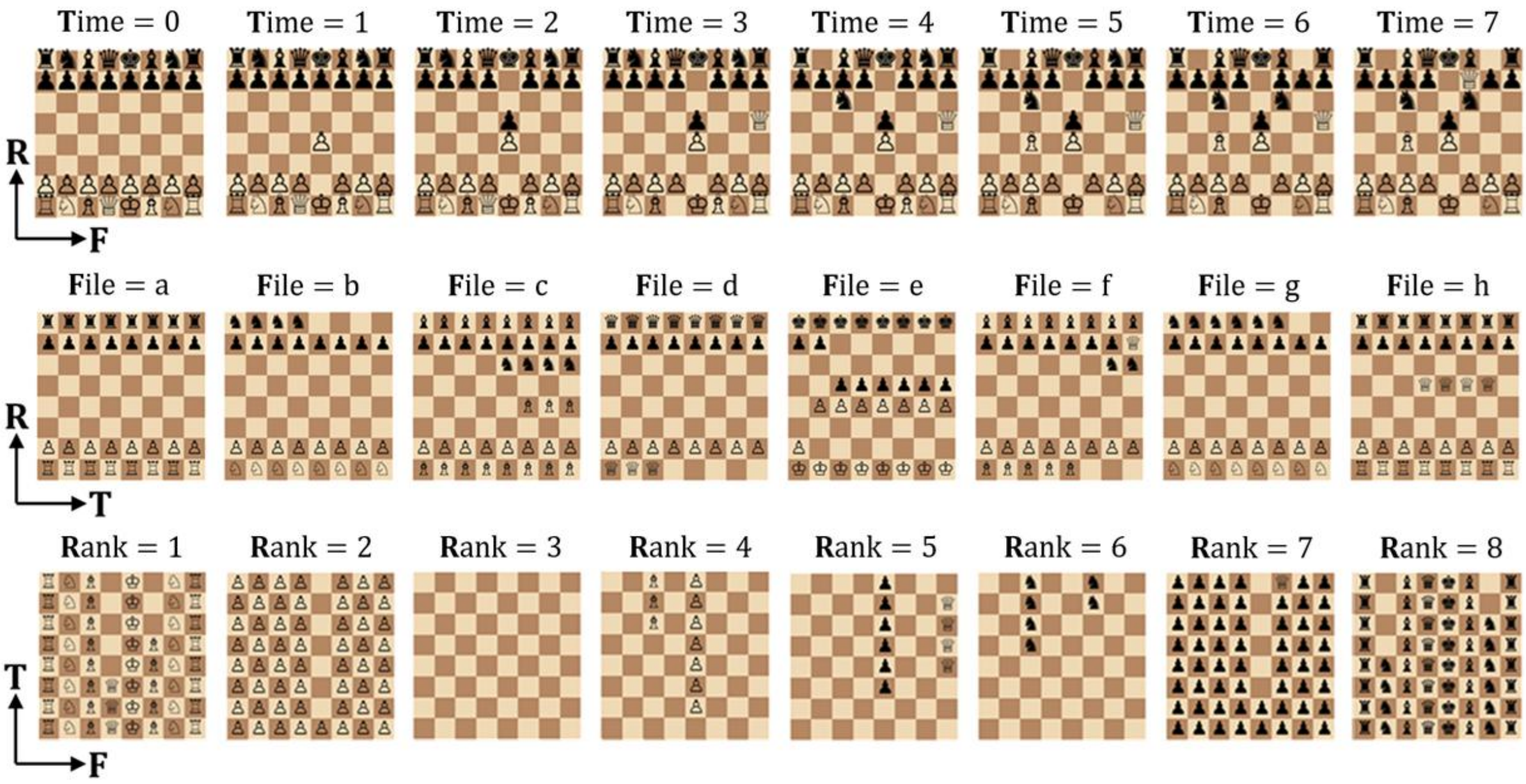

**Figure 2.** *Motivating example of the OSDH using a chess game*: The same sequence of board positions is reorganized into three ordered projections: the conventional temporal projection (top), file projections (middle), and rank projections (bottom). Each projection preserves the underlying observations while inducing a different ordering that may expose complementary structural dependencies suitable for recurrent modeling.

This motivating example captures the central intuition behind OSDH. A single phenomenon may simultaneously admit multiple meaningful ordered projections. Each projection preserves the underlying observations while inducing a different sequential organization of the data. The next section generalizes this intuition beyond chess and develops the mathematical framework for constructing such ordered projections from arbitrary multidimensional datasets.

## 3. Mathematical Foundations of OSDH

### 3.1. Problem formulation

The motivating chess example presented in the previous section illustrates a broader computational principle that is independent of the application domain. While chess naturally admits multiple ordered representations through its temporal and spatial coordinates, many real-world datasets possess analogous ordered structures that can be exploited for learning. This section formalizes the proposed OSDH and derives the corresponding multi-channel learning paradigm from first principles.

Consider an observed world $W$ to be described by a dataset $\mathcal{D}$, which is a collection of $N$ observations

$$\mathcal{D} = \{o_1, o_2, \dots, o_N\},$$

where each observation is represented as

$$o_i = (t_i, x_{i1}, x_{i2}, \dots, x_{im}, y_i).$$

Here, $t_i$ denotes the temporal index associated with observation $o_i$; $x_{ij}$ denotes the value of the $j$-th descriptive coordinate; $y_i$ denotes the target variable (or label) associated with the observation.

We assume that every observation possesses a unique temporal identifier, $t_i \neq t_j, i \neq j$, which establishes a canonical ordering of observations within the original dataset. This assumption is naturally satisfied in many sequential domains where observations correspond to successive events, measurements, or system states. More importantly, as will become clear later, the uniqueness of the temporal index guarantees deterministic construction of alternative ordered projections whenever identical values occur in non-temporal coordinates.

Therefore, the complete description of the “observed-and-to-be-modelled” world is the tuple:

$$W = \langle T, X_1, X_2, \dots, X_m, Y \rangle,$$

where $T$ denotes the temporal coordinate, $X_1, \dots, X_m$ denote the remaining descriptive coordinates characterizing each observation, and $Y$ is a target attribute.

The objective of this and similar machine learning tasks is normally to learn a predictive model: $F: (T, X_1, X_2, \dots, X_m) \rightarrow Y$, that estimates $\hat{y}_i = F(t_i, x_{i1}, \dots, x_{im})$.

However, unlike conventional learning approaches, which typically exploit only the temporal ordering induced by $T$, the proposed OSDH framework assumes that every admissible ordered coordinate $T,\ X_1, \dots, X_m$ may expose complementary sequential statistical dependencies. The central objective of OSDH is therefore not merely to learn the mapping $F$, but to learn it by jointly exploiting structural knowledge extracted independently from multiple ordered projections of the same observation set.

### 3.1. Problem formulation

The motivating chess example presented in the previous section illustrates a broader computational principle that is independent of the application domain. While chess naturally admits multiple ordered representations through its temporal and spatial coordinates, many real-world datasets possess analogous ordered structures that can be exploited for learning. This section formalizes the proposed OSDH and derives the corresponding multi-channel learning paradigm from first principles.

Consider an observed world $W$described by a dataset:

$$\mathcal{D} = \{o_1, o_2, \dots, o_N\},$$

where each observation is represented as:

$$o_i = (t_i, x_{i1}, x_{i2}, \dots, x_{im}, y_i).$$

Here, $t_i$ denotes the temporal coordinate associated with observation $o_i$; $x_{ij}$ denotes the value of the $j$-th descriptive coordinate; and $y_i$ denotes the target variable (or label).

We assume that every observation possesses a unique temporal identifier,

$$t_i \neq t_j, i \neq j,$$

which establishes a canonical ordering of observations within the original dataset. This assumption is naturally satisfied in many sequential domains where observations correspond to successive events, measurements, or system states. More importantly, as will become clear later, the uniqueness of the temporal coordinate guarantees deterministic construction of alternative ordered projections whenever identical values occur in non-temporal coordinates.

Accordingly, the observed world is represented by the tuple:

$$W = \langle T, X_1, X_2, \dots, X_m, Y \rangle,$$

where $T$ denotes the temporal coordinate, $X_1, \dots, X_m$ denote the remaining descriptive coordinates, and $Y$ is the target attribute to be predicted.

The objective is to learn a predictive model

$$F: (T, X_1, X_2, \dots, X_m) \rightarrow Y,$$

which estimates

$$\hat{y}_i = F(t_i, x_{i1}, \dots, x_{im}).$$

In conventional supervised learning, the temporal ordering induced by $T$ is typically regarded as the only sequential structure available for recurrent modeling. The proposed OSDH framework adopts a broader perspective. It assumes that every *admissible ordered coordinate* (see next subsection):

$$T,\ X_1,\ X_2,\ \dots,\ X_m$$

may induce its own ordered view of the same observation set, potentially exposing statistical dependency patterns that are not observable under the conventional temporal ordering.

Consequently, the objective of OSDH is not merely to learn the mapping $F$, but to learn it by integrating structural knowledge extracted independently from multiple ordered projections of the same world.

### 3.2. Ordered coordinates and structural dependencies

The central concept of the proposed framework is that of an *admissible ordered coordinate* (or simply *ordered coordinate*).

**Definition 1** (Admissible ordered coordinate).

A coordinate is called *admissible ordered coordinate* if its domain admits a meaningful ordering such that sorting observations according to that coordinate (using temporal order to resolve ties when necessary) produces a deterministic observation sequence suitable for recurrent modeling.

Formally, let

$$X_k = \{x_{1k}, x_{2k}, \dots, x_{Nk}\}$$

denote the values of coordinate $X_k$. If a total ordering $\prec_k$ can be defined over the domain of $X_k$, then sorting the observations according to this ordering produces the sequence

$$o_{\pi_k(1)}, o_{\pi_k(2)}, \dots, o_{\pi_k(N)},$$

where $\pi_k$ denotes the permutation ($\pi_k: \{1, \dots, N\} \rightarrow \{1, \dots, N\}$) induced by sorting the observations with respect to coordinate $X_k$. Whenever several observations possess identical values of $X_k$, their relative order is determined by their unique temporal coordinates, thereby guaranteeing a deterministic projection.

Importantly, the existence of an ordering alone does not imply that the resulting sequence contains useful predictive information. Many orderings may be mathematically valid while carrying little or no statistical regularity relevant to the learning task.

The proposed framework therefore introduces a stronger notion.

**Definition 2** (Ordered structural dependency).

An *ordered structural dependency* exists along an ordered coordinate if neighboring observations in the induced ordering exhibit statistically meaningful dependency patterns that can be exploited for prediction or representation learning.

Unlike temporal causality, ordered structural dependency does not imply a physical cause-and-effect relationship between neighboring observations. It merely states that the ordering exposes statistical regularities that recurrent computation may exploit. Temporal causality therefore represents one important manifestation of ordered structural dependency, but not its defining characteristic. Analogous dependency patterns may arise from spatial organization, hierarchical organization, geometric progression, or numerous domain-specific ordered coordinates.

These definitions provide the foundation for the OSDH, introduced in Section 1 and explained in the following subsection. OSDH does not seek to replace temporal modeling, but rather to generalize it by recognizing temporal dependency as one particular instance of a broader family of ordered structural dependencies.

### 3.3. Ordered structural dependency hypothesis (OSDH)

The conceptual motivation and intuition underlying OSDH were introduced in Section 1 and illustrated through the motivating chess example in Section 2. We now restate the hypothesis in the formal setting established in the preceding subsections.

OSDH: Let $X_k$ be an admissible ordered coordinate as defined in Definition 1. After sorting the observations according to $X_k$ (using the temporal coordinate $T$ to resolve ties whenever necessary), the resulting ordered projection may expose statistical dependency patterns that are not observable under the conventional temporal ordering. Such dependencies can be modeled using recurrent computation regardless of whether the ordering corresponds to physical time or to another admissible ordered coordinate.

OSDH is an existential hypothesis. It does not assert that every admissible ordered coordinate necessarily contains informative structural dependencies. Rather, it identifies every admissible ordered coordinate as a legitimate candidate for recurrent modeling whose usefulness must ultimately be established empirically. Likewise, OSDH does not redefine temporal causality; instead, it regards temporal dependency as one particular realization of a broader class of ordered structural dependencies.

The remaining subsections derive the proposed multi-channel learning framework directly from OSDH.

## 3.4. Construction of ordered projections

OSDH provides the conceptual motivation for exploiting multiple ordered representations of the same dataset. The next question is therefore purely algorithmic: *How can these ordered representations be constructed in a deterministic and reproducible manner*?

For every admissible ordered coordinate $X_k$ of the dataset $\mathcal{D}$, defined in Section 3.1, a new projection dataset $\mathcal{D}^{(k)}$ is constructed whose observations are ordered according to coordinate $X_k$.

Unlike conventional sequential learning, where only the temporal coordinate induces a sequence, OSDH assumes generation of one sequence for every admissible ordered coordinate, including time itself. Consequently, the original dataset gives rise to $m + 1$ ordered datasets:

$$\{\mathcal{D}^{(T)}, \mathcal{D}^{(1)}, \mathcal{D}^{(2)}, \dots, \mathcal{D}^{(m)}\},$$

where:

- $\mathcal{D}^{(T)}$ denotes the conventional ordered projection induced by the temporal coordinate $T$;
- $\mathcal{D}^{(k)}$ denotes the ordered projection induced by coordinate $X_k$, obtained by applying the corresponding permutation $\pi_k$ to $\mathcal{D}$.

Each projection contains exactly the same observations in different order. This distinction is fundamental as OSDH does not generate new data. Instead, it generates multiple ordered representations of the same world, each potentially exposing different structural dependency patterns.

### 3.4.1. The dual role of time

Although OSDH treats every admissible ordered coordinate identically during recurrent learning, the temporal coordinate retains a second role during data preparation.

Specifically, time serves two independent purposes within the framework.

In its *learning role*, the temporal coordinate defines the conventional recurrent channel, $\mathcal{D}^{(T)}$, which models classical temporal dependencies exactly as in traditional sequential learning.

In its *preprocessing role*, the temporal coordinate simultaneously provides the canonical identifier of observations. Whenever identical values occur in a non-temporal coordinate, $x_{ik} = x_{jk}$, the corresponding observations are ordered according to their temporal indices, $t_i < t_j$. Consequently, every ordered projection becomes deterministic.

It is important to emphasize that this second role does not imply that time is considered more informative than other coordinates. Rather, it guarantees reproducibility of the projection construction process, while the learning model itself remains entirely symmetric with respect to all admissible ordered coordinates.

### 3.4.2. Ordered projection operator

The construction of each projection can now be defined formally.

Let $\Pi_k(\mathcal{D})$ denote the projection operator associated with coordinate $X_k$. The operator performs two steps:

- Step 1. Sort all observations according to coordinate $X_k$.
- Step 2. Whenever two observations possess identical coordinate values, order them according to their temporal identifiers as mentioned above.

The resulting ordered projection is therefore:

$$\mathcal{D}^{(k)} = \Pi_k(\mathcal{D}) = \left(o_{\pi_k(1)}, o_{\pi_k(2)}, \dots, o_{\pi_k(N)}\right),$$

where the permutation $\pi_k$ is determined lexicographically by $(x_{ik}, t_i)$.

All projection datasets contain exactly the same observations and therefore have identical cardinality. They differ exclusively in the ordering of observations.

This definition guarantees that every admissible coordinate produces exactly one deterministic sequence. Notice that the temporal projection $\mathcal{D}^{(T)}$ is simply a special case in which sorting is performed directly by temporal index.

**Algorithm 1.** Construction of ordered projection datasets.

Input: Original dataset $\mathcal{D} = \{(t_i, x_{i1}, \dots, x_{im}, y_i)\}_{i=1}^{N}$.

Output: Projection datasets $\{\mathcal{D}^{(T)}, \mathcal{D}^{(1)}, \dots, \mathcal{D}^{(m)}\}$.

For every admissible coordinate $C \in \{T, X_1, \dots, X_m\}$ apply the projection operator $\Pi_C$ on $\mathcal{D}$, i.e., perform:

1. Sort observations according to coordinate $C$.
2. If equal coordinate values occur, preserve temporal order using $t$.
3. Store the resulting ordered projection dataset $\mathcal{D}^{(C)}$.

Return: All projection datasets. ■

**Algorithm 1** represents the fundamental preprocessing operation of OSDH. Its computational role is intentionally simple: the algorithm never modifies observations, synthesizes features, or performs dimensionality reduction. Instead, it constructs multiple deterministic ordered views of the same dataset. Each ordered projection preserves the complete information content of the original dataset while inducing a different neighborhood structure among the observations. Consequently, different projection datasets may reveal different statistical regularities, allowing subsequent learning algorithms to discover complementary structural dependencies that remain hidden under conventional temporal ordering alone.

## 3.5. Independent Structural Expertise Principle

The OSDH establishes that a single world may admit several ordered representations, each potentially exposing different statistical dependency patterns. A natural question therefore arises: *Should these ordered projections be learned jointly from the outset, or should each projection first be allowed to develop its own specialized representation*?

To adopt the latter viewpoint, we formulate the Independent Structural Expertise Principle (ISEP) as follows: *Each admissible ordered projection of a dataset should be regarded as an independent source of structural knowledge. A dedicated learning model is assigned to every ordered projection and trained independently to discover the statistical regularities exposed by that particular ordering. Knowledge acquired by different projection-specific models is integrated only after the individual models have converged.*

ISEP follows naturally from the philosophy underlying OSDH. Since different ordered projections reveal different neighborhood structures among the same observations, forcing all projections to interact during early learning may encourage premature co-adaptation and reduce the diversity of the learned representations. Instead, each projection is treated as an independent learning task whose objective is to fully exploit the structural dependencies characteristic of its own ordered view.

Consequently, OSDH can be interpreted as a society of independent structural experts. Every expert observes the same underlying world but through a different ordered representation. Although all experts ultimately address the same prediction task, each specializes in discovering structural regularities that may remain hidden from experts operating on other projections.

**Figure 3** summarizes the two-stage learning paradigm implied by ISEP.

This interpretation differs fundamentally from conventional ensemble learning. Classical ensembles usually obtain diversity through different initialization, different subsets of training data, or different learning algorithms. Under ISEP, every expert receives exactly the same observations and solves the same learning problem. Diversity is induced exclusively by the different ordered projections of the same observations rather than by stochastic training effects or heterogeneous datasets. Similarly, ISEP differs from standard multi-branch neural architectures, where branches are commonly optimized jointly through a shared loss function. In contrast, the proposed framework deliberately postpones interaction between projection-specific experts until each expert has independently acquired its own structural competence.

The purpose of ISEP is therefore to maximize representational diversity before integration. The subsequent fusion stage combines independently acquired structural competencies rather than jointly optimized intermediate representations.

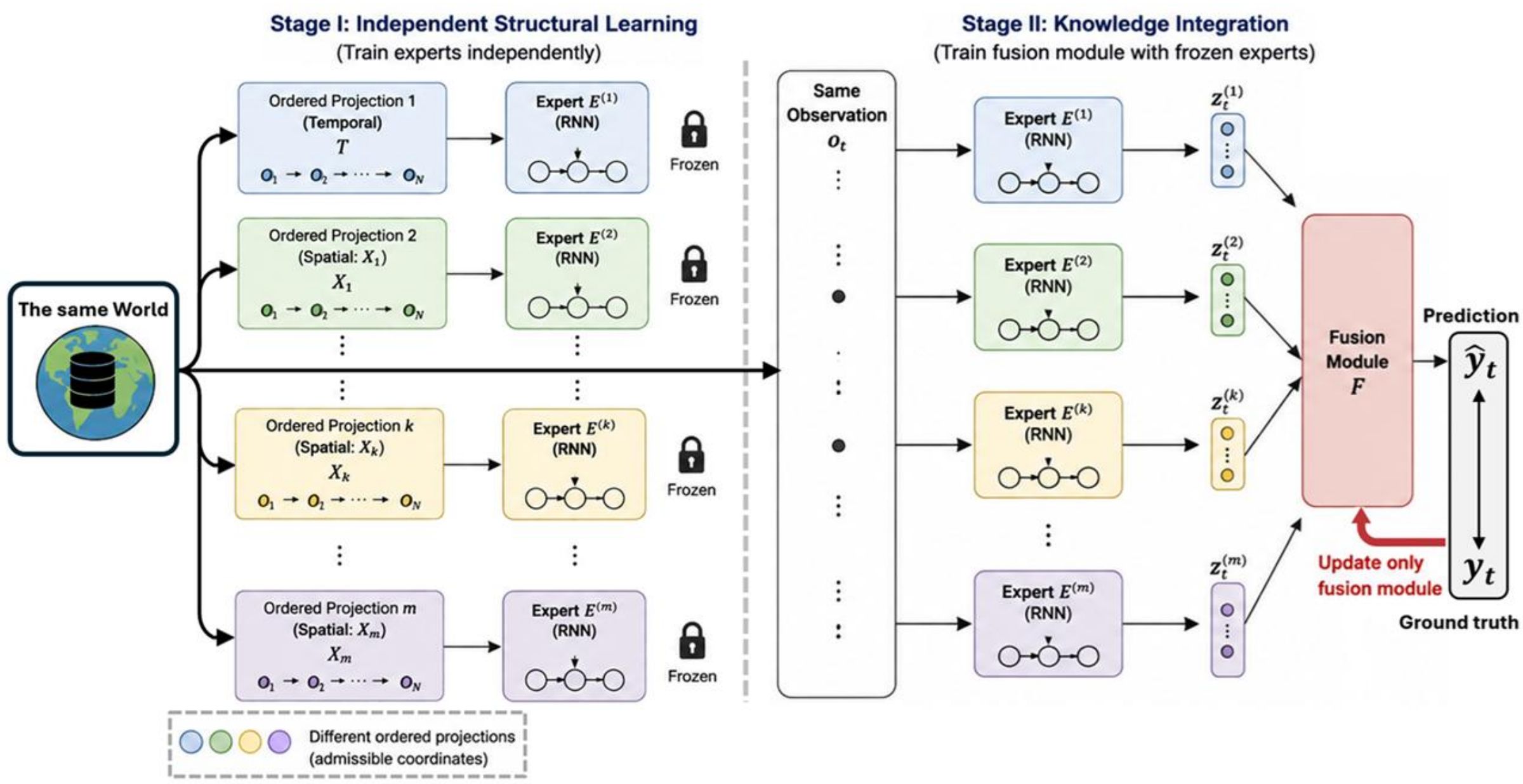

**Figure 3.** *Independent Structural Expertise Principle*: The same world is projected onto multiple admissible ordered coordinates. Each ordered projection is assigned to an independent recurrent expert that learns projection-specific structural dependencies. After independent training, the experts are frozen and their outputs are integrated by a learnable fusion module.

## 3.6. Projection-specific sequential learning

Once the ordered projection datasets have been constructed according to **Algorithm 1**, each ordered projection defines an independent supervised learning problem.

For every admissible coordinate $C \in \{T, X_1, \dots, X_m\}$, **Algorithm 1** produces the ordered projection dataset $\mathcal{D}^{(C)}$ defined in Subsection 3.4. Each dataset is assigned to its own projection-specific structural expert $E^{(C)}$. In the experimental realization presented in this paper, every structural expert is implemented as a RNN.

Although the experimental realization presented in this paper employs RNNs as projection-specific experts, the proposed framework is intentionally more general. OSDH requires only that the learning model be capable of exploiting statistical dependencies within ordered observations. Consequently, alternative sequential architectures, including gated recurrent networks, state-space models, or other sequence-processing mechanisms, may equally serve as structural experts without changing the underlying OSDH formulation.

For a recurrent implementation, let $o_t^{(C)}$ denote the observation occupying position $t$in the ordered projection dataset $\mathcal{D}^{(C)}$. Its input features are denoted by $x_t^{(C)}$, while $y_t^{(C)}$ denotes the corresponding target value. Each structural expert computes the hidden-state recursion

$$h_t^{(C)} = f^{(C)}\left(h_{t-1}^{(C)}, x_t^{(C)}\right),$$

where:

- $x_t^{(C)}$ denotes the input feature vector of the observation occupying position $t$ in projection $\mathcal{D}^{(C)}$;
- $h_t^{(C)}$ denotes the corresponding latent representation;
- $f^{(C)}$ denotes the recurrent transition function.

Depending on the implementation, $f^{(C)}$ may correspond to a vanilla RNN, a Long Short-Term Memory (LSTM) network , a Gated Recurrent Unit (GRU), or another recurrent architecture.

The parameters of every structural expert are optimized independently by minimizing the supervised objective

$$\theta_C^* = \arg\min_{\theta_C} \sum_{t=1}^{N} \mathcal{L}\left(g_{\theta_C}(h_t^{(C)}), y_t^{(C)}\right),$$

where:

- $\theta_C$ denotes the parameters of structural expert $E^{(C)}$;
- $g_{\theta_C}$ denotes its prediction function;
- $\mathcal{L}(\cdot)$ is the task-specific loss function.

In accordance with ISEP, no information, gradients, or intermediate representations are exchanged between projection-specific experts during this optimization stage. Each expert therefore converges independently toward the parameter values that best capture the statistical regularities exposed by its own ordered projection.

Only after all experts have completed training are their parameters frozen. From this point onward, every structural expert becomes a fixed projection-specific encoder capable of producing structural representations for previously unseen observations.

The two-stage separation between knowledge acquisition and knowledge integration is a direct consequence of ISEP. Independent experts first maximize the structural information extractable from their own ordered projections. Their learned competencies are subsequently combined by a learnable fusion module into a unified predictive model.

## 3.7. Knowledge integration through learnable fusion

As discussed in the previous subsection, ISEP separates learning into two conceptually distinct stages: (i) independent acquisition of projection-specific structural knowledge and (ii) learning how to integrate the independently acquired (and subsequently frozen) structural knowledge into a unified prediction.

### 3.7.1. Two-stage learning procedure

The complete OSDH framework consists of two consecutive optimization problems.

*Stage I: Independent Structural Learning*

Each ordered projection $\mathcal{D}^{(C)}$ is assigned to its own structural expert $E^{(C)}$. Every expert is trained independently according to the optimization problem introduced in the previous subsection. After convergence, the learned parameters $\theta_C^*$ are fixed. No further updates of projection-specific models are performed. At this point each expert represents a specialized structural competency associated with one ordered projection of the world.

*Stage II: Knowledge Integration*

Although each expert was trained using its own ordered projection, after training every expert becomes a fixed projection-specific encoder that can generate a structural representation for any observation appearing within its corresponding projection.

After all projection-specific experts have been trained and frozen, each ordered projection is processed once more. During this inference pass, every observation $o_i$ is encountered in the structural context induced by the corresponding projection, allowing expert $E^{(C)}$ to produce a projection-specific representation:

$$z_i^{(C)} = E^{(C)}(o_i).$$

The collection $\left\{z_i^{(T)}, z_i^{(1)}, \dots, z_i^{(m)}\right\}$ forms a joint structural representation of observation $o_i$.

The projection-specific representations are then provided to the learnable fusion model $F_\phi$, which computes the final prediction:

$$\hat{y}_i = F_\phi\left(z_i^{(T)}, z_i^{(1)}, \dots, z_i^{(m)}\right),$$

where $\phi$ denotes the parameters of the fusion model.

During this second learning stage, the projection-specific experts remain fixed and serve exclusively as feature extractors. Consequently, only the fusion parameters are optimized. The corresponding optimization problem is:

$$\phi^* = \arg\min_{\phi} \sum_{i=1}^{N} \mathcal{L}\left(F_\phi\left(z_i^{(T)}, z_i^{(1)}, \dots, z_i^{(m)}\right), y_i\right),$$

while $\theta_C = \theta_C^*, \forall\, C \in \{T, X_1, \dots, X_m\}$.

Since the projection-specific experts remain frozen throughout Stage II, optimization updates only the fusion parameters $\phi$. Thus, the second learning stage is devoted exclusively to integrating the complementary structural knowledge already acquired by the independently trained experts, without modifying the expertise itself.

### 3.7.2. Conceptual interpretation

The proposed learning procedure admits an intuitive interpretation. Consider a complex world observed simultaneously by several independent specialists. Each specialist studies exactly the same world but through a different structural perspective. One specialist analyzes the temporal projection, while the remaining specialists analyze alternative ordered projections induced by the admissible coordinates of the world. Each expert develops its own specialized competence independently. Only after these competencies have matured are the experts invited to collaborate. A separate decision-maker then learns how much confidence should be assigned to each specialist for solving the final prediction task. Importantly, the specialists themselves are not retrained during collaboration, while only the decision-maker improves. Consequently, knowledge acquisition and knowledge integration remain conceptually separated in accordance with OSDH philosophy and ISEP.

### 3.7.3. Why not train everything jointly?

An obvious alternative would be to optimize all projection-specific experts and the fusion network simultaneously using a single global loss function. Although technically feasible, such optimization follows a different learning philosophy. Joint optimization encourages projection-specific experts to co-adapt throughout training. As a result, different experts may gradually converge toward similar internal representations, thereby reducing the structural diversity that OSDH seeks to exploit. Independent optimization encourages every expert to specialize exclusively in the statistical regularities exposed by its own ordered projection. Only after these complementary competencies have been acquired does the fusion network learn how to combine them effectively. From the OSDH perspective, diversity of structural expertise is regarded as a desirable property rather than an optimization artifact. For this reason, the proposed framework adopts independent optimization as its default learning strategy.

Therefore, the two-stage learning procedure is not merely a convenient implementation strategy. Rather, it is a direct consequence of the theoretical assumptions underlying OSDH. If different ordered projections are assumed to expose complementary structural knowledge, then preserving the independence of their learning processes becomes a natural methodological choice. The subsequent fusion stage therefore addresses a qualitatively different problem: not discovering new structural dependencies but learning how to integrate already acquired structural expertise into a coherent global prediction.

This design philosophy distinguishes OSDH from conventional end-to-end optimization strategies, where diversity among internal representations emerges only implicitly through the optimization process. Under OSDH, structural diversity is treated as an explicit design objective and is therefore preserved throughout the learning procedure.

## 3.8. Summarized OSDH learning pipeline

The preceding subsections introduced the theoretical principles underlying the proposed framework. OSDH motivated the construction of multiple ordered projections of the same dataset, ISEP justified independent learning of projection-specific models, and the two-stage optimization procedure separated structural knowledge acquisition from knowledge integration. These components are summarized in **Algorithm 2**, which presents the complete OSDH learning pipeline.

**Algorithm 2.** Complete OSDH learning framework

**Input:**

- World $W = \langle T, X_1, \dots, X_m, Y \rangle$;
- Dataset $\mathcal{D} = \{o_1, o_2, \dots, o_N\}$ derived from observations of $W$.

*Stage I: Construction of ordered projections*

Use **Algorithm 1** to construct $\{\mathcal{D}^{(T)}, \mathcal{D}^{(1)}, \dots, \mathcal{D}^{(m)}\}$.

*Stage II: Independent structural learning (ISEP)*

For every ordered projection $\mathcal{D}^{(C)}, C \in \{T, X_1, \dots, X_m\}$ perform independently:

1. Initialize projection-specific expert $E^{(C)}$.

2. Train $E^{(C)}$ using only $\mathcal{D}^{(C)}$ until convergence.
3. Freeze the learned parameters $\theta_C^*$.

*Stage III: Knowledge integration*

Initialize fusion model $F_\phi$.

Repeat until convergence:

For each ordered projection $\mathcal{D}^{(C)}, C \in \{T, X_1, \dots, X_m\}$:

- Process the entire ordered projection $\mathcal{D}^{(C)}$ using the corresponding frozen expert $E^{(C)}$.
- Obtain the projection-specific representations $\left\{z_1^{(C)}, z_2^{(C)}, \dots, z_N^{(C)}\right\}$.

For every training observation $o_i \in \mathcal{D}$:

- Collect its projection-specific representations $z_i^{(T)}, z_i^{(1)}, \dots, z_i^{(m)}$.
- Compute the integrated prediction $\hat{y}_i = F_\phi\left(z_i^{(T)}, z_i^{(1)}, \dots, z_i^{(m)}\right)$.
- Compute the task loss $\mathcal{L}(\hat{y}_i, y_i)$.

Update only the parameters of the fusion model.

**Output:**

A trained OSDH model consisting of:

- independently optimized structural experts $\left\{E^{(T)}, E^{(1)}, \dots, E^{(m)}\right\}$,
- a trained fusion model $F_\phi$,

which together produce each final prediction: $\hat{y}_i = F_\phi(z_i^{(T)}, z_i^{(1)}, \dots, z_i^{(m)})$. ■

**Algorithm 2** emphasizes that the proposed framework extends conventional sequential learning by constructing multiple ordered representations of the same dataset rather than by modifications of recurrent computation itself. The novelty of the framework lies in three complementary ideas:

1. OSDH, which generalizes sequential learning from temporal order to arbitrary admissible ordered coordinates.
2. ISEP, which assigns an independently trained expert to each ordered projection in order to maximize projection-specific structural specialization.
3. Knowledge integration through learnable fusion, which combines independently acquired structural competencies into a unified predictive model without modifying the learned experts.

Consequently, the proposed framework should not be viewed as a new recurrent architecture or a new recurrent cell. Rather, it constitutes a general computational paradigm that extends recurrent learning from a single privileged ordering to multiple complementary ordered representations of the same underlying world.

## 4. Structural Evolution RNNs (SE-RNNs): A Concrete OSDH Realization

The theoretical framework developed in Section 3 is intentionally independent of the particular sequence-processing architecture used to model ordered projections. Any model capable of learning from ordered sequences could, in principle, serve as a projection-specific expert within the OSDH framework. In this work, we instantiate OSDH using RNNs and refer to the resulting architecture as Structural Evolution RNNs (SE-RNNs), which have been already mentioned in Section 1 and conceptually (and partially) visualized in **Figure 1** and **Figure 3**. The name emphasizes that each recurrent model learns the evolution of structural patterns observed along one admissible ordered projection of the same world. Physical time therefore becomes one particular realization of structural evolution rather than its defining case. Unlike Section 3, which established the general computational paradigm, the present section describes the specific neural architecture used throughout the experimental study.

### 4.1. Structural evolution

Within the OSDH framework, *structural evolution* (SE) denotes the ordered progression of observations along a particular admissible coordinate. When the projection coordinate is the temporal coordinate $T$, SE coincides with classical temporal dynamics. For all remaining admissible coordinates, SE describes ordered changes of the same observations with respect to the corresponding structural organization of the world. Consequently, SE should be understood as an architectural concept describing what each projection-specific expert intends to model, rather than as a separate theoretical hypothesis. OSDH explains why such ordered projections may contain useful dependencies, whereas SE characterizes the projection-specific sequential patterns learned by an individual expert.

### 4.2. Overall architecture

The overall architecture has been conceptually illustrated in **Figures 1** and **3**. The computational workflow of SE-RNN follows the OSDH learning pipeline established in Section 3. Rather than introducing additional learning principles, the architecture specifies how the theoretical components are instantiated in the experimental system. Multiple ordered projections are processed by independently trained recurrent experts whose learned representations are subsequently combined with a conventional feedforward backbone through a learnable fusion module.

### 4.3. Projection-specific structural encoders

As established in Section 3.6, each ordered projection is assigned to an independently trained structural expert. In the implementation presented here, every expert is realized as an RNN, giving rise to the name SE-RNN. No modifications of conventional recurrent computation are introduced; the contribution lies in organizing multiple recurrent experts according to the OSDH/ISEP framework rather than in proposing new recurrent architecture.

### 4.4. Feedforward decision backbone

In addition to the recurrent experts, the implemented architecture contains a conventional feedforward neural network (FFNN) operating directly on the original feature representation. Its role differs fundamentally from that of the recurrent experts. While recurrent experts encode SE associated with individual ordered projections, the feedforward backbone captures relationships directly available from the original feature representation without relying on sequential processing. The backbone therefore provides complementary non-sequential information that can be combined with the structural representations extracted by the recurrent experts.

### 4.5. Learnable fusion

The latent representations produced by the frozen SE-RNN experts, together with the representation generated by the FFNN backbone, are supplied to a learnable fusion module. Unlike the projection-specific experts, whose parameters remain fixed after Stage I of learning, the fusion component is optimized during the second training stage described in Section 3. Its objective is therefore not to discover additional structural dependencies but rather to learn how to combine complementary structural competencies already acquired by the independently trained experts.

The present implementation employs a feedforward fusion network, although alternative fusion mechanisms could equally be incorporated within the same framework.

### 4.6. Relationship between OSDH and SE-RNN

The relationship between the proposed concepts can now be summarized as follows:

- OSDH provides the theoretical hypothesis that multiple admissible ordered projections may reveal complementary structural dependencies.
- ISEP specifies the learning methodology by requiring projection-specific experts to acquire their structural competencies independently.
- SE denotes the ordered structural patterns modeled by an individual projection-specific expert.
- SE-RNN is the concrete neural implementation of OSDH and ISEP in which projection-specific experts are realized using RNNs.

Consequently, the proposed contribution should be viewed hierarchically. OSDH constitutes the general computational principle, ISEP specifies the corresponding learning strategy, and SE-RNN represents one concrete realization of these concepts using RNNs. Alternative sequence-processing architectures may replace the recurrent experts without affecting the theoretical foundations established in Section 3.

## 5. Experimental Validation of the OSDH Framework Using SE-RNNs

### 5.1. Experimental setup

The objective of the experimental study is not to benchmark a new recurrent architecture against existing state-of-the-art models. Rather, the experiments are intended to provide a proof-of-concept validation of the theoretical framework introduced in Sections 3 and 4. Specifically, they investigate whether the use of multiple ordered projections of the same dataset, combined with independent projection-specific learning and subsequent knowledge integration, can improve predictive performance over a conventional feedforward baseline.

To instantiate the proposed framework, we employ the SE-RNN architecture introduced in Section 4. SE-RNN represents one concrete realization of OSDH in which every ordered projection is processed by an independent recurrent structural expert, while the resulting projection-specific representations are integrated through a learnable fusion module together with a conventional feedforward backbone.

The experiments were intentionally conducted on synthetic datasets. This choice provides complete control over the structural relationships embedded in the data and allows the influence of different ordered projections to be studied under well-defined conditions. Rather than attempting exhaustive benchmarking across numerous application domains, the goal is to isolate the contribution of the proposed learning paradigm under datasets exhibiting different degrees of hidden structural complexity.

The implementation (Python/PyTorch), dataset generation procedures, complete source code, hyperparameter settings, and additional experimental results are available in the Supplementary Material (available online in: https://ai.it.jyu.fi/experiments/SE-RNNs/).

*Experimental datasets*

Three synthetic datasets were generated to represent qualitatively different levels of structural complexity among three input variables $X = (x_1, x_2, x_3)$ and the target variable $y$.

In the following dataset definitions, $\varepsilon$ denotes additive Gaussian noise, and $\varepsilon \sim \mathcal{N}(\mu, \sigma^2)$ indicates that it is sampled from a normal distribution with mean $\mu$ and variance $\sigma^2$.

*Dataset 1* contains smooth nonlinear relationships generated from trigonometric and hyperbolic interactions:

$$y = \sin(x_1 \cdot x_2) + \cos(x_2 \cdot x_3) + \tanh(x_1 - x_3) + \varepsilon, \text{ where } \varepsilon \sim \mathcal{N}(0, 0.1).$$

The dataset contains continuous cross-feature interactions that are moderately nonlinear while remaining highly structured.

*Dataset 2* introduces substantially more complex nonlinear relationships:

$$y = \cos(x_1) \cdot x_2 - \sin(x_2 \cdot x_3) + 0.1 \cdot \varepsilon,$$

combining multiplicative and oscillatory dependencies that create less obvious structural regularities.

*Dataset 3* serves as a deliberately simple (and partially adversarial) control dataset:

$$y = 1.5 \cdot x_1 - 2 \cdot x_2 + 0.5 \cdot x_3 + \varepsilon, \text{ where } \varepsilon \sim \mathcal{N}(0, 0.05^2).$$

Its target depends almost entirely on a linear combination of independent features, providing little opportunity for additional ordered projections to reveal complementary structural dependencies.

Taken together, the three datasets represent increasing levels of structural complexity. Dataset 1 evaluates moderately nonlinear structural relationships, Dataset 2 represents substantially more challenging hidden interactions, whereas Dataset 3 provides a setting in which the proposed OSDH framework is expected to offer only limited additional benefit over conventional feedforward learning.

*Experimental realization*

The experimental realization follows the two-stage learning procedure introduced in Section 3.

For every dataset, the original observations were transformed into multiple ordered projection datasets according to **Algorithm 1**. Each ordered projection was assigned to an independent recurrent structural expert implemented as an RNN. After independent training, the projection-specific experts were frozen and used as structural encoders. Their learned representations were subsequently combined with the representation produced by a conventional FFNN through a learnable fusion module.

The present implementation should be viewed as one practical realization of the OSDH framework rather than its only possible implementation. As discussed in Sections 3 and 4, OSDH itself is independent of the particular sequence-processing architecture used for projection-specific learning.

*Training protocol*

To investigate the contribution of different architectural components, three modules were trained:

- the feedforward backbone (FFNN),
- the projection-specific recurrent structural experts (RNNs),
- the learnable fusion module.

For each component, two training regimes were considered:

- *Weak:* 25-30 training epochs;
- *Strong:* 100 training epochs.

This produced eight combinations of component strengths, ranging from "Weak-Weak-Weak" to "Strong-Strong-Strong" (regarding the FFNN-RNN-Fusion settings).

For every experiment, predictive performance was evaluated using Mean Squared Error (MSE). The reported results include the baseline FFNN performance, the corresponding SE-RNN performance, the relative improvement with respect to the FFNN baseline, and qualitative observations regarding the behavior of the different architectural components.

Complete implementation details, including neural network architectures, optimizer settings, learning rates, batch sizes, activation functions, hidden dimensions, and training schedules, are provided in the Supplementary Material.

## 5.2. Results on Dataset 1

Dataset 1 represents a learning problem with moderately complex nonlinear interactions among the input features. According to the OSDH hypothesis, different ordered projections of such a dataset may expose complementary statistical dependencies that are only partially accessible from the original feature representation. The experiments therefore evaluate whether the implemented SE-RNN realization can exploit these complementary projections to improve predictive performance.

**Table 1:** SE-RNN performance (MSE) evaluation on Dataset 1 with moderately complex hidden relationships.

| Run | FFNN | RNNs | Integrator | FFNN | SE-RNN | (%) Relative improvement | Notes on performance |
|---|---|---|---|---|---|---|---|
| 1 | WEAK | WEAK | WEAK | 0.0380 | 0.0198 | 47.9 | All components are weak; integration still yields substantial improvement |
| 2 | STRONG | WEAK | WEAK | 0.0158 | 0.0142 | 10.5 | Strong FFNN; weak structural experts limit additional improvement. |
| 3 | WEAK | STRONG | WEAK | 0.0380 | 0.0121 | 68.2 | Strong structural experts capture complementary ordered dependencies |
| 4 | WEAK | WEAK | STRONG | 0.0380 | 0.0118 | 69.0 | Strong fusion effectively exploits even weak structural representations |
| 5 | STRONG | STRONG | WEAK | 0.0158 | 0.0121 | 23.8 | Strong experts; weak fusion limits knowledge integration |
| 6 | WEAK | STRONG | STRONG | 0.0380 | **0.0105** | **72.4** | Strong structural experts and fusion compensate for a weak FFNN backbone |
| 7 | STRONG | WEAK | STRONG | 0.0158 | 0.0125 | 21.3 | Strong FFNN and fusion maintain good performance despite weak structural experts |
| 8 | STRONG | STRONG | STRONG | 0.0158 | 0.0111 | 30.0 | All components strong; best integrated result |

Several observations can be drawn from the results presented in **Table 1**.

First, the SE-RNN implementation consistently outperforms the standalone FFNN baseline in all experimental configurations. The magnitude of the improvement depends on the relative training quality of the individual architectural components, but the overall trend remains stable across all eight experiments.

Second, improvements become particularly pronounced when the projection-specific recurrent experts receive sufficient training (Runs 3, 5, 6 and 8). This observation is consistent with the underlying OSDH assumption that independently learned ordered projections may capture complementary structural information unavailable to the feedforward backbone alone.

Third, the fusion module plays an important role in converting projection-specific representations into improved predictions. Even when individual recurrent experts are only weakly trained (Run 4), a well-trained fusion model is able to utilize the available structural representations effectively, substantially reducing prediction error.

Finally, the strongest overall performance is achieved when all architectural components are well trained (Run 8), indicating that the feedforward backbone, projection-specific structural experts, and learnable fusion module provide complementary rather than competing sources of predictive information.

Dataset 1, therefore, provides empirical evidence that the proposed OSDH learning strategy can exploit multiple ordered projections to improve prediction accuracy on problems exhibiting moderately complex hidden structural relationships.

## 5.3. Results on Dataset 2

Dataset 2 was intentionally designed to contain substantially stronger nonlinear interactions and more intricate hidden dependencies than Dataset 1. Such data provides a more demanding test of the proposed framework because the predictive relationships are less directly observable from the original feature representation.

According to OSDH, increasing structural complexity should increase the likelihood that different ordered projections expose complementary dependency patterns. Consequently, larger improvements from projection-specific learning may be expected than for Dataset 1. The experimental results presented in **Table 2** support this expectation.

**Table 2:** SE-RNN performance (MSE) evaluation on Dataset 2 with highly complex hidden relationships.

| Run | FFNN | RNNs | Integrator | FFNN | SE-RNN | (%) Relative improvement | Notes on performance |
|---|---|---|---|---|---|---|---|
| 1 | WEAK | WEAK | WEAK | 0.3709 | 0.2402 | 35.2 | All weak; SE-RNN shows moderate gain |
| 2 | STRONG | WEAK | WEAK | 0.0278 | 0.0881 | -216.9 | Strong FFNN; weak structural experts and weak fusion fail to exploit the strong baseline |
| 3 | WEAK | STRONG | WEAK | 0.3709 | 0.0470 | 87.3 | Strong structural experts capture complementary ordered dependencies |
| 4 | WEAK | WEAK | STRONG | 0.3709 | 0.1239 | 66.6 | Strong fusion effectively exploits weak structural representations |
| 5 | STRONG | STRONG | WEAK | 0.0278 | 0.0377 | -35.5 | Strong experts; weak fusion limits knowledge integration |
| 6 | WEAK | STRONG | STRONG | 0.3709 | 0.0282 | **92.4** | Strong structural experts and fusion compensate for a weak FFNN backbone |
| 7 | STRONG | WEAK | STRONG | 0.0278 | **0.0217** | 22.1 | Strong FFNN + fusion; performs well even with weak RNNs |
| 8 | STRONG | STRONG | STRONG | 0.0278 | 0.0220 | 21.0 | All components strong; good integrated result |

One may see that, when both the recurrent experts and the fusion module receive sufficient training (Runs 6–8), the implemented SE-RNN consistently achieves substantially lower prediction errors than the corresponding feedforward baseline. In particular, Run 6 demonstrates that strong projection-specific experts combined with an effective fusion model can compensate for a relatively weak feedforward backbone, suggesting that the structural representations extracted from multiple ordered projections contain highly informative predictive information.

Several experiments (Runs 2 and 5) exhibit negative values of relative improvement. These results should be interpreted carefully. In both cases the standalone FFNN already provides exceptionally strong predictions, whereas either the recurrent experts or the fusion module remain comparatively undertrained. Under such conditions the integrated model cannot yet fully exploit the additional structural representations, resulting in slightly inferior performance. This behavior reflects the current implementation and optimization strategy rather than contradicting the underlying OSDH hypothesis.

More importantly, when all architectural components are adequately trained (Run 8), the integrated model again outperforms the standalone FFNN. This indicates that the complementary information extracted from multiple ordered projections remains beneficial even for considerably more complex nonlinear learning problems.

Dataset 2 demonstrates that the advantages of the proposed framework become more pronounced as hidden structural relationships increase in complexity. The results are therefore consistent with the central premise of OSDH: complementary ordered projections may reveal predictive dependencies that cannot be fully exploited through a single representation of the data.

### 5.4. Results on Dataset 3: A simple (adversarial) learning problem

To complement the experiments on Datasets 1 and 2, which contain increasingly complex hidden structural dependencies, we evaluated the proposed SE-RNN implementation on Dataset 3 introduced in Subsection 5.1. This dataset was intentionally designed as a simple (and partially adversarial) benchmark in which the target depends primarily on a linear combination of independent input features with only small additive Gaussian noise. Consequently, conventional feedforward learning is expected to be nearly sufficient, leaving little opportunity for additional ordered projections to reveal complementary structural dependencies. From the perspective of OSDH, this dataset therefore represents an intentionally unfavorable scenario for the proposed framework.

The same eight experimental configurations used for the previous datasets were evaluated, varying the training strength (WEAK/STRONG) of the FFNN, projection-specific RNN experts, and the fusion module. The results are summarized in **Table 3**.

**Table 3:** SE-RNN performance evaluation (MSE) on Dataset 3 with predominantly linear relationships.

| Run | FFNN | RNNs | Integrator | FFNN | SE-RNN | (%) Relative improvement | Notes on performance |
|---|---|---|---|---|---|---|---|
| 1 | WEAK | WEAK | WEAK | 0.0028 | 0.0038 | -34.36 | All components weak; added structural modeling provides little benefit |
| 2 | STRONG | WEAK | WEAK | 0.0026 | 0.0039 | -49.08 | Strong FFNN; weak structural experts and fusion provide little complementary information |
| 3 | WEAK | STRONG | WEAK | 0.0028 | 0.0026 | 7.40 | Strong structural experts provide modest complementary information despite weak fusion |
| 4 | WEAK | WEAK | STRONG | 0.0028 | 0.0029 | -2.61 | Strong fusion alone cannot compensate for weak structural experts |
| 5 | STRONG | STRONG | WEAK | 0.0026 | 0.0026 | 0.23 | Strong experts; weak fusion limits effective integration |
| 6 | WEAK | STRONG | STRONG | 0.0028 | **0.0025** | **10.59** | Strong structural experts and fusion yield the best integrated performance |
| 7 | STRONG | WEAK | STRONG | 0.0026 | 0.0034 | -31.03 | Weak structural experts limit the effectiveness of a strong fusion module |
| 8 | STRONG | STRONG | STRONG | 0.0026 | 0.0029 | -10.95 | Near-optimal FFNN leaves little opportunity for complementary structural information |

Unlike Datasets 1 and 2, the performance differences between the standalone FFNN and the integrated SE-RNN architecture are generally small. This outcome is consistent with the construction of the dataset itself. Since the target function contains little hidden structural complexity, there is correspondingly little additional information that projection-specific experts can contribute to. Several observations nevertheless deserve attention.

First, the implemented SE-RNN remains competitive with the FFNN baseline across all experimental settings. Although several configurations exhibit slight degradations in predictive performance, these degradations remain modest, indicating that the proposed architecture does not become unstable even when multiple ordered projections contribute little useful information.

Second, configurations employing well-trained projection-specific experts (Runs 3 and 6) still produce measurable improvements over the corresponding weak FFNN baseline. This suggests that even relatively simple datasets may contain weak structural regularities that become exploitable through multiple ordered projections.

Third, when the FFNN backbone is already highly optimized (Runs 2, 7 and 8), the additional projection-specific information becomes largely redundant. Under these circumstances the fusion stage occasionally introduces slight overfitting, producing marginally higher prediction errors than the standalone FFNN. Such behavior is expected when the original representation already captures nearly all predictive information.

Therefore, Dataset 3 serves as an important negative control for the proposed framework. Rather than demonstrating large improvements under all circumstances, the results show behavior that is fully consistent with OSDH: the benefit of multiple ordered projections depends on the existence of meaningful structural dependencies that can be exposed by alternative orderings. When such complementary dependencies are largely absent, the proposed framework naturally provides only limited additional benefit over conventional feedforward learning.

### 5.5. Comparative analysis across the three datasets

Taken together, the three experimental datasets reveal a consistent relationship between structural complexity and the effectiveness of the proposed architecture.

Dataset 3, whose target function is predominantly linear, provides only limited opportunity for complementary ordered projections to contribute additional predictive information. Consequently, the implemented SE-RNN performs similarly to the standalone FFNN, with only modest improvements or slight degradations depending on the particular training configuration.

Dataset 1 introduces moderate nonlinear interactions among the input features. Under these conditions, independently learned projection-specific representations consistently improve predictive accuracy, indicating that different ordered projections expose complementary structural dependencies that can be successfully integrated during the second learning stage.

Dataset 2 further increases structural complexity through stronger nonlinear and multiplicative interactions. Here the benefits of projection-specific learning become even more pronounced, particularly when both the recurrent experts and the fusion module receive sufficient training. The improvements support the OSDH hypothesis that richer hidden structural relationships increase the value of considering multiple admissible ordered projections.

Several general observations emerge from the complete experimental study:

- Influence of structural complexity: The predictive advantage of the proposed architecture increases as hidden structural relationships become more complex. This trend is consistent across all three datasets.
- Importance of independent projection-specific learning: Well-trained structural experts consistently provide more informative representations than weakly trained experts, confirming the practical importance of the independent learning strategy introduced in Section 3.
- Role of the fusion stage: The fusion model contributes primarily by learning how to combine complementary projection-specific representations. Its effectiveness therefore depends on the quality of the structural representations generated during the first learning stage.
- Robustness of the implementation: Even on the intentionally unfavorable Dataset 3, the integrated architecture remains competitive with the standalone FFNN, indicating that the introduction of additional ordered projections does not inherently destabilize the learning process.

Therefore, the experiments provide consistent empirical support for the practical realization of OSDH through the proposed SE-RNN architecture. While the current implementation represents only one possible realization of the theoretical framework, the observed performance trends closely follow the predictions derived from the theoretical principles established in Section 3.

### 5.6. Overall experimental assessment

The experiments consistently support the theoretical motivation developed in Sections 3 and 4. Across datasets with progressively different structural characteristics, the proposed SE-RNN implementation behaves as predicted by the OSDH framework.

On Datasets 1 and 2, where the target variables depend on increasingly complex hidden interactions among the input attributes, the projection-specific structural experts learned complementary ordered representations that improved predictive performance after knowledge integration. The observed improvements become particularly pronounced for Dataset 2, indicating that the proposed framework is most beneficial when multiple ordered projections expose structural dependencies that are not readily captured from the original feature representation alone. In contrast, Dataset 3 intentionally provides little opportunity for complementary ordered projections to reveal additional information. Under these conditions, the implemented architecture remains competitive with the FFNN baseline but produces only limited improvement, which is consistent with our theoretical expectation that OSDH provides the greatest benefit when meaningful projection-specific structural dependencies exist.

These experiments should be interpreted as a proof-of-concept validation of the proposed computational paradigm rather than as an exhaustive benchmark. Their primary contribution is to demonstrate that organizing learning around multiple admissible ordered projections constitutes a viable and practically useful extension of conventional recurrent learning.

### 5.7. Computational considerations

An important practical property of the proposed implementation is its modular organization. During Stage I, each projection-specific expert is trained independently using only its corresponding ordered projection. Because no information is exchanged among experts during this stage, their optimization can be performed fully in parallel whenever appropriate computational resources are available. The dominant computational cost therefore scales primarily with the cost of training a single expert rather than requiring sequential optimization across all projections.

After the experts have been trained and frozen, Stage II optimizes only the fusion model. Since the projection-specific representations remain fixed throughout this stage, optimization becomes substantially simpler than end-to-end training of complete architecture. The fusion network operates on already computed structural representations rather than repeatedly updating all recurrent parameters.

Consequently, the implemented SE-RNN architecture exhibits two practical advantages:

- First, the independent optimization strategy naturally supports distributed and parallel computation.
- Second, separating structural representation learning from knowledge integration substantially simplifies the second optimization stage and reduces the risk of unstable interactions among projection-specific experts.

Although the computational cost increases with the number of ordered projections considered, the modular nature of the framework makes this increase manageable and readily parallelizable. Future implementations employing alternative sequence-processing architectures may further improve computational efficiency while preserving the theoretical principles introduced in Section 3.

### 5.8. Discussion, limitations, and future directions

The present study intentionally focuses on validating the conceptual foundations of OSDH through one concrete implementation based on RNNs. Several limitations should therefore be acknowledged.

First, the experimental evaluation was conducted using synthetic datasets specifically designed to exhibit different degrees of hidden structural complexity. This design was intentional, since synthetic data provide complete control over the underlying dependency structure and therefore allow the proposed computational principles to be evaluated under well-understood conditions. Rather than maximizing benchmark performance, the objective was to isolate the influence of multiple ordered projections on predictive accuracy. A natural question, however, concerns the relatively small number of datasets used in the experimental study. The objective of the experiments was not exhaustive benchmarking across many unrelated datasets, but rather controlled evaluation across qualitatively different structural regimes. The three synthetic datasets were intentionally designed to span a broad spectrum of conditions relevant to the proposed framework: Dataset 3 represents problems with weak or nearly absent hidden structural dependencies, Dataset 1 represents moderately complex nonlinear interactions, and Dataset 2 represents highly entangled nonlinear relationships with stronger cross-feature coupling. Together, these datasets cover a progression from simple to structurally rich learning problems, allowing the experiments to test both the boundary conditions under which OSDH provides little additional benefit and the conditions under which multiple ordered projections become increasingly informative. In this sense, the diversity of the chosen datasets is more important than their number: the experimental design was intended to probe different classes of structural complexity rather than repeatedly evaluate similar datasets.

Second, the current implementation considers manually specified ordered projections derived from domain knowledge. OSDH itself does not require this assumption. Future research may investigate automatic projection discovery using optimization-based ordering procedures, graph decomposition techniques, differentiable sorting operators, attention mechanisms, or meta-learning approaches capable of identifying informative ordered representations directly from data.

Third, the present work employs RNNs as projection-specific experts because they provide the most direct realization of the proposed framework. However, the theoretical formulation developed in Section 3 is intentionally architecture-independent. Alternative sequence-processing models (including gated recurrent networks, state-space models, transformer-based architectures, or future sequence-learning mechanisms) could replace the recurrent experts without modifying the underlying OSDH formulation.

Another important direction concerns the fusion stage. The current implementation employs a relatively simple learnable fusion network whose purpose is to integrate independently learned structural representations. More sophisticated fusion strategies, including attention-based, probabilistic, graph-based, or adaptive mixture-of-experts approaches, may further improve performance while remaining fully consistent with the proposed two-stage learning paradigm.

Finally, although the present experiments demonstrate encouraging and consistent performance across datasets of different structural complexity, broader empirical validation remains an important direction for future work. Evaluation on real-world datasets, larger-scale statistical studies involving repeated experimental runs, confidence intervals, significance testing, and comparisons with additional sequence-learning architectures will provide a more comprehensive assessment of the practical applicability of the proposed framework.

The experimental results indicate that treating ordering as a design variable rather than restricting recurrence to temporal sequences constitutes a promising direction for future sequence-learning research. OSDH provides the corresponding theoretical framework, while the SE-RNN architecture presented in this paper demonstrates one practical realization of this broader computational paradigm.

## 6. Related Work

The proposed framework combines three complementary ideas: the OSDH, which treats ordering as a general computational principle rather than a property of time alone; the ISEP, which advocates independent learning on multiple ordered projections; and SE-RNNs, which constitute one concrete neural realization of these principles. Accordingly, the related literature is reviewed from several complementary perspectives: sequence modeling and recurrent architectures, the conceptual distinction between ordering and time, multi-dimensional sequence processing, multi-view and representation learning, modular neural architectures, and hybrid fusion-based models. This organization reflects the theoretical emphasis of the present work, where SE-RNN is viewed as an implementation of the broader OSDH framework rather than its primary contribution.

### 6.1. Sequence modeling and recurrent architectures

RNNs constitute one of the foundational paradigms for learning from ordered data. By recursively updating an internal hidden state as successive observations are processed, RNNs naturally capture sequential dependencies of arbitrary length [1]. Early recurrent architectures, however, suffered from vanishing and exploding gradient problems that limited their ability to learn long-range dependencies [10]. These limitations motivated the development of gated recurrent architectures, including LSTM networks [11] and GRUs [12], which substantially improved optimization stability through explicit memory and gating mechanisms.

Subsequent developments extended recurrent learning in several directions, including bidirectional recurrent networks [13], deep and stacked recurrent architectures [14], and reservoir-based approaches such as Echo State Networks [15]. These models have been successfully applied to numerous application domains, including language modeling and machine translation [16-17], speech recognition [18], time-series forecasting [19], and system identification [20-21]. Comprehensive surveys summarize both their expressive power and their inductive bias toward sequential processing [22, 3].

Although Transformer architectures [6] have become the dominant sequence-processing paradigm for many application domains, recurrent models continue to offer important advantages in settings requiring explicit state evolution, streaming computation, or lightweight deployment. Consequently, recurrent learning remains an active area of research despite the rapid development of attention-based architectures.

The present work does not propose a new recurrent cell, memory mechanism, or optimization algorithm. Instead, it adopts conventional recurrent models as projection-specific sequence processors within a broader computational framework. Consequently, the novelty of the proposed approach lies not in modifying recurrence itself, but in redefining what constitutes a meaningful sequence to which recurrent computation may be applied.

## 6.2. Order versus time in sequence learning

A fundamental conceptual question concerns the relationship between sequential ordering and physical time. In many practical applications, RNNs are introduced as models for temporal dynamics because their inputs naturally arrive as time-indexed observations. However, from a computational perspective, the recurrence mechanism itself depends only on the existence of an ordered sequence rather than on any intrinsic notion of time. As emphasized in theoretical analyses of recurrent networks, the hidden-state update operates over an ordered index, irrespective of whether that index represents time, spatial position, linguistic structure, or another ordered variable [22, 14]. This distinction has become increasingly relevant as sequence-learning research has expanded well beyond classical time-series analysis. This view aligns also with philosophical and cognitive science interpretations in which ordering, causality, and dependency need not coincide with physical time [23]. Contemporary sequence-processing architectures, including recurrent networks, Transformers, and more recent state-space models, are routinely applied to language, biological sequences, program code, symbolic reasoning, and numerous other domains in which the sequential index does not correspond to physical time. These diverse applications reinforce the general view that sequence modeling concerns learning statistical dependencies over ordered observations, irrespective of whether the ordering is temporal, spatial, symbolic, or otherwise defined.

The OSDH builds directly upon this general interpretation of sequence learning. Rather than assuming that temporal order is the unique or privileged organization of data, OSDH treats any admissible ordered coordinate capable of exposing statistically meaningful dependencies as a legitimate sequence-processing axis. Temporal ordering therefore represents one important special case, but not the only possible realization of ordered structural dependencies. Importantly, the proposed framework does not claim that arbitrary orderings are universally beneficial. Instead, it hypothesizes that multiple admissible orderings of the same underlying observations may reveal complementary dependency structures that remain inaccessible when learning is restricted to a single sequential organization. This hypothesis distinguishes OSDH from conventional sequence-learning approaches, which generally assume a single predefined ordering and focus primarily on improving the architecture used to process that sequence rather than reconsidering how the sequence itself is constructed.

## 6.3. Multi-dimensional and structural recurrent architectures

Several research directions have attempted to extend recurrent computation beyond a single temporal dimension. One of the earliest examples is the Multi-Dimensional RNN (MDRNN), which generalizes recurrence across multiple spatial dimensions by propagating hidden states along different axes of structured grids [18]. MDRNNs have been successfully applied to handwriting recognition and image processing, where spatial ordering is naturally defined.

More recently, recurrent processing has been combined with graph-based representations. Structural RNNs [24] model interactions among entities connected through predefined spatio-temporal graphs, while recurrent graph neural networks integrate graph message passing with sequential state updates [25-26]. Related developments include graph sequence models that jointly learn relational and temporal dependencies in dynamic graphs [27].

Although these architectures successfully exploit structured domains, they fundamentally assume that the underlying structural organization (grids, graphs, or relational topologies) is already available or can be explicitly constructed. The proposed OSDH framework addresses a different problem. Rather than extending recurrence over predefined structures, OSDH generates multiple ordered projections directly from the original dataset, allowing several independent structural experts to discover complementary statistical regularities associated with different admissible orderings.

Consequently, the proposed SE-RNN implementation differs from MDRNNs and graph-recurrent models in two important respects. First, recurrence is performed independently within each ordered projection rather than jointly across a multidimensional computational graph. Second, the independently learned representations are integrated only after projection-specific learning has completed, following the ISEP principle introduced in Section 3. This separation between independent structural learning and subsequent knowledge integration distinguishes the proposed framework from existing multidimensional recurrent architectures.

### 6.4. Multi-view learning, multi-modal learning, and ensemble representation learning

Multi-view learning investigates learning problems in which multiple representations of the same object are available [28]. Classical approaches include co-training, co-regularization, and shared latent-space learning, while modern deep-learning formulations learn complementary feature representations jointly across views [29].

A closely related research direction is multi-modal learning, where heterogeneous data sources such as text, images, audio, or sensor streams are integrated into a common predictive model [30-31]. Recent advances increasingly rely on cross-modal attention mechanisms and shared representation learning to align heterogeneous modalities [32].

Another relevant line of research concerns ensemble representation learning and modular neural architectures, in which multiple independently trained components are combined through learned aggregation mechanisms [33-34]. Such Mixture-of-Experts architectures demonstrate that independently specialized subnetworks can improve predictive performance when their competencies are appropriately coordinated.

Despite these similarities, OSDH differs fundamentally from conventional multi-view and multi-modal learning. Existing approaches assume that multiple views or modalities already exist as separate data sources. In contrast, OSDH generates multiple complementary representations internally by constructing different admissible ordered projections of the same dataset. The resulting representations therefore differ not because different information sources are available, but because the same observations are organized according to different ordering principles.

Similarly, although the proposed fusion stage resembles ensemble aggregation or Mixture-of-Experts architectures, its objective is different. The projection-specific experts are not expected to become alternative predictors of the same task. Instead, each expert specializes in modeling structural dependencies exposed by one particular ordered projection, while the fusion model learns how to combine these complementary structural representations into a unified prediction.

### 6.5. Tensor methods and higher-order representations

Tensor-based learning provides another important perspective on modeling complex interactions in multidimensional data. Tensor decompositions such as CANDECOMP/PARAFAC and Tucker decomposition provide compact representations of higher-order relationships and have been successfully applied to recommender systems, knowledge representation, signal processing, and neural network compression [35-36].

More recent studies have investigated tensorized neural architectures, tensor recurrent units, and multilinear representation learning to improve parameter efficiency while preserving expressive power [37]. Tensor methods have also found increasing applications in deep learning through tensor factorization of neural weights and tensor-based representation learning [38].

Although these approaches effectively capture higher-order interactions, they typically represent such relationships implicitly through multilinear algebraic structures. In contrast, the OSDH framework exposes complementary structural dependencies explicitly by constructing multiple admissible ordered projections of the same dataset and learning them independently through projection-specific experts. Rather than replacing tensor-based representations, the proposed framework offers an orthogonal computational perspective in which higher-order relationships emerge from complementary ordered traversals instead of tensor decompositions.

### 6.6. Hybrid architectures, attention mechanisms, and foundation sequence models

Hybrid neural architectures combining recurrent networks with complementary processing modules have been extensively investigated. CNN-RNN hybrids employ convolutional layers for local feature extraction followed by recurrent modeling of sequential dependencies [39] (Shi et al., 2015). Similar combinations have been successfully applied to speech processing, video understanding, biomedical signal analysis, and many other domains.

Attention mechanisms, originally introduced for neural machine translation [17], enable models to dynamically emphasize the most relevant elements within a sequence and substantially improve the modeling of long-range dependencies. The subsequent introduction of the Transformer architecture [6] replaced recurrence entirely with self-attention, establishing a new paradigm for sequence modeling.

Since then, transformer-based architectures have expanded far beyond natural language processing into computer vision, time-series analysis, scientific computing, and multimodal learning [40-41]. Despite their impressive empirical performance, these models continue to process a single ordered representation of the input unless multiple views are explicitly provided.

The relationship between OSDH and these architectures is complementary rather than competitive. OSDH is not tied to recurrent computation itself. As discussed in Sections 3 and 4, any sequence-processing architecture capable of exploiting ordered dependencies—including recurrent networks, Transformers, state-space models, or future sequence learners—could serve as a projection-specific expert. The present work adopts recurrent networks because they provide the most direct and mathematically transparent realization of the proposed computational principle.

Consequently, the novelty of the present work lies neither in proposing another recurrent architecture nor in replacing attention-based models, but in introducing a new mechanism for generating multiple complementary ordered representations upon which any suitable sequence-processing architecture may subsequently operate.

### 6.7. Positioning of OSDH, ISEP, and SE-RNN relative to prior work

The reviewed literature demonstrates remarkable advances in sequence modeling, representation learning, structural learning, and neural integration. Nevertheless, existing approaches generally fall into one of several categories:

- Single-order sequence learning, where one predefined ordering (typically temporal) determines the sequential structure processed by the model.
- Structure-dependent sequence learning, where recurrence or message passing is defined over known grids, graphs, spatial layouts, or relational structures.
- Multi-view or multimodal learning, where multiple externally available representations are fused into a common predictive model.
- Mixture-of-Experts and modular neural architectures, where multiple experts are combined through learned routing or aggregation mechanisms.

The proposed framework differs from each of these directions in a fundamental way.

First, OSDH introduces the hypothesis that the same dataset may admit several admissible ordered projections, each exposing different statistical dependencies. Unlike previous work, these projections are not assumed to exist naturally but are generated from alternative orderings of the original observations.

Second, ISEP proposes that each ordered projection should be learned independently before any interaction among experts occurs. This explicit separation between projection-specific knowledge acquisition and subsequent knowledge integration distinguishes the proposed learning strategy from conventional end-to-end multi-branch optimization.

Finally, SE-RNN represents one concrete realization of these principles using RNNs. Importantly, SE-RNN is not presented as a new recurrent architecture. Rather, it demonstrates how conventional recurrent networks can be reorganized according to OSDH and ISEP in order to exploit multiple complementary ordered representations of the same underlying world.

Viewed from this perspective, the primary contribution of the present work is not architectural innovation at the level of individual neural components, but the introduction of a new computational paradigm for sequence learning in which order itself becomes a design variable rather than an inherent property of temporal data.

## 7. Conclusions

This work advances a broader view of sequence learning by introducing OSDH. The proposed perspective considers ordering itself as a modeling variable that may be deliberately selected to expose informative structural regularities within the same observations. From this viewpoint, temporal ordering is one important instance among many possible admissible orderings capable of supporting sequential learning. Consequently, the proposed framework extends the scope of sequence-processing methods beyond conventional time-indexed data while remaining fully compatible with existing sequential learning models.

Building upon this hypothesis, the paper proposed ISEP, which separates learning into two consecutive stages. During the first stage, projection-specific experts independently acquire structural knowledge from their corresponding ordered projections without exchanging gradients or intermediate representations. During the second stage, the learned structural competencies are integrated through a dedicated fusion model while the projection-specific experts remain frozen. This separation between knowledge acquisition and knowledge integration provides a conceptually simple and modular learning strategy that is independent of the particular sequence-processing architecture employed.

To demonstrate the practical applicability of these principles, the paper introduced SE-RNNs as one concrete realization of the proposed framework. Within this implementation, RNNs serve as projection-specific structural experts, while the notion of SE characterizes the ordered dependency patterns modeled along each admissible projection. Accordingly, the contribution of SE-RNNs lies in extending the application domain of recurrent computation to multiple admissible ordered projections, while leaving the underlying recurrent mechanisms unchanged. Consequently, OSDH remains applicable to alternative sequence-processing architectures beyond recurrent networks.

The experimental study provided proof-of-concept validation of the proposed framework using three synthetic datasets exhibiting substantially different levels of structural complexity. The results consistently demonstrate that the benefits of the implemented SE-RNN architecture become increasingly pronounced as the structural complexity of hidden dependencies within the data increases. For datasets containing complex nonlinear interactions, independently learned projection-specific experts produced complementary structural representations whose integration yielded substantial improvements over a standalone feedforward network. Conversely, for a deliberately simple dataset in which the original feature representation already captured nearly all predictive information, the advantages of additional ordered projections naturally diminished. These observations are fully consistent with the theoretical motivation underlying OSDH: complementary ordered projections become valuable precisely when they expose structural dependencies that remain hidden within a single representation of the data.

The experiments should be interpreted as validation of the proposed computational principles rather than as an attempt to establish state-of-the-art predictive performance on benchmark datasets. The objective of the synthetic experimental design was to isolate the influence of multiple ordered projections under controlled conditions while evaluating the fundamental assumptions underlying OSDH and ISEP. The encouraging empirical results indicate that these assumptions deserve further validation on more complex real-world datasets and application domains.

Several directions for future research naturally follow from the proposed framework. An important next step is the application of OSDH-based architectures to practical domains in which multiple meaningful orderings coexist, including industrial process monitoring, spatio-temporal analytics, biological systems, multimodal data, and graph-structured information. Equally important is the development of methods capable of automatically discovering informative ordered projections rather than relying on domain knowledge. Finally, since OSDH itself is independent of any particular sequence-processing model, future research may investigate implementations based on modern state-space models, Transformer architectures, graph neural networks, or their hybrids.

Taken together, the theoretical developments and experimental results presented in this paper support a new perspective on sequence learning in which ordering is treated as an explicit modeling variable rather than as a fixed property of data. OSDH and ISEP establish a general computational framework for exploiting complementary ordered representations of the same observations, while SE-RNNs demonstrate that these principles can be instantiated using existing recurrent networks without modifying their internal computational mechanisms. We hope that this perspective will stimulate further research on sequence-learning architectures capable of exploiting multiple complementary structural organizations of data.